\pdfoutput=1
\documentclass[11pt]{article}

\usepackage[]{acl}
\usepackage{enumitem}

\usepackage{contour}
\usepackage{xcolor}
\usepackage{times}
\usepackage{latexsym}
\usepackage{graphicx}
\usepackage{subcaption}
\usepackage{mwe}
\usepackage{xcolor}
\usepackage{booktabs}
\usepackage{amsmath,amsfonts,amssymb,amsthm}
\usepackage{multicol}\usepackage{xcolor}         % colors
\usepackage{diagbox}
\contourlength{0.8pt}
\usepackage{makecell}

\newcommand{\xhdr}[1]{\vspace{1mm}\noindent{{\bf #1.}}}

\definecolor{amaranth}{HTML}{E52B50}
\definecolor{garnet}{HTML}{733635}
\definecolor{burgundy}{HTML}{800020}
\definecolor{midnightgreenn}{rgb}{0.0, 0.29, 0.33}
\definecolor{navyblue}{rgb}{0.0, 0.0, 0.5}

\usepackage[T1]{fontenc}
\usepackage[utf8]{inputenc} % The default since 2018
\usepackage[english]{babel}
\newcommand\dedicationfont{\fontfamily{qcr}\itshape\mdseries\selectfont}
\newcommand{\wc}{{\dedicationfont \textcolor{navyblue}{Word Confusion}}\xspace}
\newcommand{\remove}[1]{}

\usepackage[T1]{fontenc}
\usepackage[utf8]{inputenc}

\usepackage{microtype}

\usepackage{inconsolata}
\usepackage{booktabs}
\title{Rethinking Word Similarity:\\Semantic Similarity through Classification Confusion}

\author{
  Kaitlyn Zhou, 
  Haishan Gao, 
  Sarah Chen, 
  Dan Edelstein,   
  Dan Jurafsky, 
  Chen Shani\\
  Stanford University\\
\texttt{\{katezhou, hsgao, sachen, danedels, jurafsky, cshani\}@stanford.edu} \\  
}

\begin{document}
\maketitle

\begin{abstract}
%Word similarity is critical in its applications to humanistic and social science tasks, such as measuring semantic changes over time, detecting biases, and making sense of contested terms. Yet, traditional similarity methods based on cosine similarity between word embeddings falls short in capturing the context-dependent, asymmetrical, polysemous nature of semantic similarity. Inspired by \citet{tversky1977features}'s cognitive model of conceptual tasks, we propose a new measure of similarity focused on extracting and compiling context-relevant features. We present \wc, a model that reframes semantic similarity in terms of feature-based \textit{classification confusion}. Specifically, we train a classifier to map contextual embeddings to word identities and use the classifier confusion (the probability of choosing a confounding word $c$ instead of the correct target word $t$) as a measure of the similarity of $c$ and $t$. Our method outperforms cosine similarity in matching human similarity judgments across several datasets (MEN, WirdSim353, and SimLex), can measure similarity using predetermined features of interest, and enables qualitative analysis of real-world data. 
Word similarity has many applications to social science and cultural analytics tasks like measuring meaning change over time and making sense of contested terms. 
Yet traditional similarity methods based on cosine similarity between word embeddings 
cannot capture the context-dependent, asymmetrical, polysemous nature of semantic similarity.
We propose a new measure of similarity,  \wc,
that reframes semantic similarity in terms of feature-based \textit{classification confusion}.
\wc is inspired by \citet{tversky1977features}'s suggestion
that similarity features be chosen dynamically.
Here we train a classifier to map contextual embeddings to word identities and use the classifier confusion (the probability of choosing a confounding word $c$ instead of the correct target word $t$) as a measure of the similarity of $c$ and $t$. The set of potential confounding words acts as the chosen features. Our method is comparable to cosine similarity in matching human similarity judgments across several datasets
(MEN, WirdSim353, and SimLex), and can measure similarity using predetermined features of interest.
We demonstrate our model's ability to make use of dynamic features by applying it to test a hypothesis about changes in the 18th C. meaning of the French word \textit{``révolution''} from {\em popular} to {\em state} action during the French Revolution. 
We hope this reimagining of semantic similarity will inspire the development of new tools that better capture the multi-faceted and dynamic nature of language, advancing the fields of computational social science and cultural analytics and beyond.
\end{abstract}
\section{Introduction}
\label{sec:intro}

% Word similarity serves as a cornerstone for various tasks, including measuring word sense change \cite{}, detecting biases in language \cite{}, and understanding contested concepts like art, nature, and market \cite{}.
Semantic similarity measures allow computational social scientists, digital humanists, and NLP practitioners to perform fine-grained synchronic and diachronic analysis on word meaning, with important applications to areas like cultural analytics and legal and historical document analysis \cite{bhattacharya2020methods, rios2012dissimilarity}. 

The cosine between two embedding vectors is the most commonly used similarity metric for textual analysis across a variety of fields 
%including the digital humanities 
\cite{johri-etal-2011-study, caliskan2017semantics, manzini-etal-2019-black, martinc-etal-2020-leveraging}. However, it neither accounts for the multi-faceted nature of similarity  \cite[inter alia]{tversky1977features, ettinger2016evaluating, zhou2022problems} nor does it align exactly with how humans perceive similarity \cite{nematzadeh2017evaluating}. Cosine similarity is dominated by a small number of rogue dimensions due to the anisotropy of contextual embedding spaces \cite{timkey2021all, ethayarajh2019contextual}, underestimates the semantic similarity of high-frequency words \citep{zhou2022problems}, is a symmetric metric that cannot capture the asymmetry of semantic relationships\footnote{For example, human similarity judgments are known to be directional; \textit{``cat''} is more similar to \textit{``animal'}' than \textit{``animal''} is to \textit{``cat''}.} \citep{vilnis2014word}, and often fails in capturing human interpretation \cite{sitikhu2019comparison}. 

Here, we propose to think about concept similarity metrics differently. We are inspired by \citet{tversky1977features}'s seminal work on similarity, presuming that humans have a rich mental representation of concepts. When faced with a particular task, like similarity assessment, we extract and compile from this rich representation only the relevant features for the required task. This formulation highlights the multi-faceted and context-dependent nature of similarity judgments \cite{evers2014revisiting}.
 
To demonstrate the potential of this new framing, we introduce a proof-of-concept: \wc, a self-supervised method that {\bf defines the semantic similarity between words according to a classifier's confusion between them}. In a nutshell, we first train a classifier to map from a word embedding to the word itself, distinguishing it from a set of distractors. At inference time, given a new embedding $e$ for a target word $t$, the probability the classifier assigns to a confound word $c$ is used as a measure of similarity of words $c$ and $t$. The set of distractor words used in training act as \textit{features}, thus, the \textbf{similarity between words is based on their feature interchangeability}. 

\begin{figure*}[h!]
    \centering
    \begin{subfigure}[]{0.47\textwidth}
        \centering
        \label{fig:word_confusion_a}
        \includegraphics[width=\textwidth]{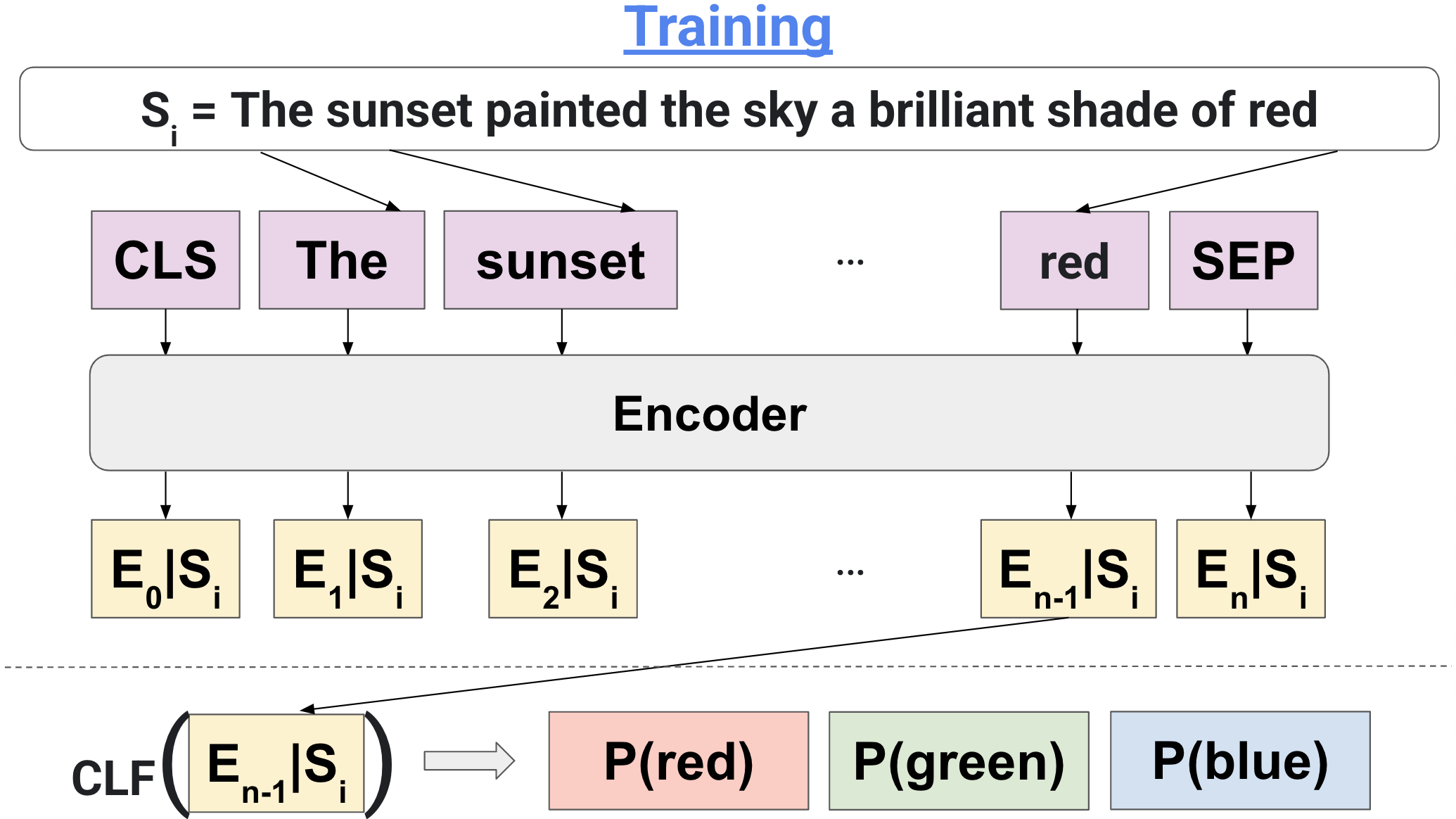}
        \caption{Training \wc: The classifier is trained in a self-supervised manner, after selecting the desired features (in this example the classes red, green blue). We extract sentences containing those 3 ``feature'' words. The input to the classifier is the contextual embedding of the class token, e.g., the BERT embedding of the word ``red'' in the sentence ``The sunset painted the sky a brilliant shade of red''. The classifier is trained to predict a class (``red'') from that contextual embedding.}
    \end{subfigure}
% \par\bigskip
\qquad
% \par\medskip
    \begin{subfigure}[]{0.47\textwidth}
        \centering
        \label{fig:word_confusion_b}
        \includegraphics[width=\textwidth]{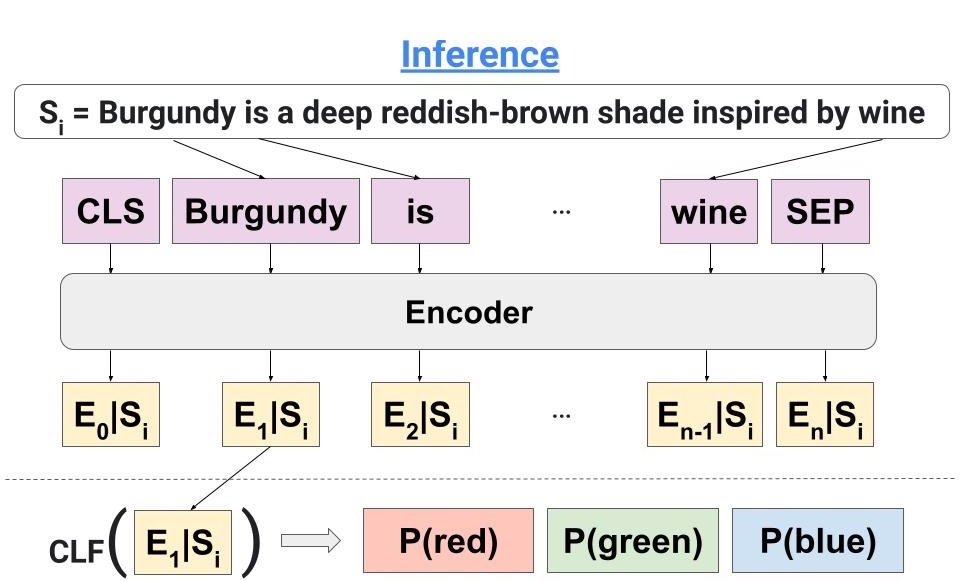}
        \caption{\wc inference: We are given the classifier and the predetermined set of classes, which will act as features, in this case red, green, blue. 
        Given a target word in a sentence, e.g., \textit{``burgundy''}, we extract its contextual embedding in that sentence $e_{\text{burgundy}}$ and compute $P(w_i|e_{\text{burgundy}})$ for each class $i$. The classifier's confusion matrix then define the similarity of the burgundy with each class. The input word can be one of the feature words (red, green, blue) or not (burgundy).}
     \end{subfigure}
\par\medskip
    \caption{{\wc}: We predetermine a set of classes for our classifier, in this case \{red, green, blue\}. This choice of classes defines the similarity features used to describe the input word. At training, we extract sentences containing the chosen class words \{red,\ green,\ blue\}. We then train the classifier to map from a BERT contextual embeddings of these words to right class \slash feature (color, in this case). At inference, we extract BERT's contextual embeddings of a target word, that may be a class (red) or may be a new word  (\textit{``burgundy''}). We then input the embedding to the classifier and use its confusion matrix to understand which primary colors burgundy is similar to.}
    \label{fig:word_confusion}
    % \par\medskip
\end{figure*}

We test \wc on standard word-similarity tasks like sentiment and grammatical gender classification and show that it is comparable to standard cosine similarity and can be more meaningful. We then apply \wc to real-world data exploration tasks in collaboration with the fourth author, who is a scholar of French literature and history. We use the {\em Archive Parlementaires} from 1789-1793 to study a long debated question in the political history of revolutionary France: how ``revolution'' went from being seen as a means of popular liberation, to becoming identified with governmental actions that often flouted such personal freedoms. 

% We do this by measuring the \wc similarity of the French word ``revolution'' to different sets of words in the French {\em Archive Parlementaires} from 1789-1793. 

\noindent In summary, our paper:
\begin{enumerate}[itemsep=0pt]
   \item Proposes a novel framing of semantic similarity, inspired by cognitive models and sensitive to the blind spots of cosine similarity. Our new formulation can learn more complex word identity boundaries than cosine similarity alone; accounts for the asymmetrical nature of semantic similarity; can be easily adapted to desired domains; and provides a more interpretable measure. 
   \item Implements a proof-of-concept of our new framing of similarity, showing it is comparable to cosine on standard semantic similarity benchmarks.\footnote{Our data and code can be found: \url{https://github.com/sally9805/word-confusion}}
   \item Applies our method to real-world data, showcasing its potential for analyzing word meaning and temporal trends.
\end{enumerate}

We hope this new formulation will spark the creation of tools for cultural analytics and computational social science that account for the multi-faceted and complex nature of semantic similarity.

\section{Introducing \wc}
\label{sec:identity_probe}

Our method begins by defining a set of words, or features. For example, we might choose the set $W=\{red,\ green,\ blue\}$ if we wanted to study similarity related to colors. These words will act as features that can be selected by the analyst to focus on a particular dimension or question.

Our process then has two phases: training and inference. In training (illustrated in part (a) of Figure \ref{fig:word_confusion})) we extract from a corpus a set of sentences containing each of these words, such as \textit{``The sunset painted the sky a brilliant shade of red''} for the word ``red''. We then use BERT to extract the contextual embeddings of these feature-words, and train a classifier to map from an embedding to its corresponding word identity. The classifier's training objective is to correctly classify the embedding to the word that corresponds to it. 

More formally, given a chosen set of word $W$ and embeddings $\{e_1, e_2 .... e_i\} \in E$ that correspond to word identities $\{w_1, w_2, ..., w_i\} \in W$, we train a logistic regression classifier on all pairs of $\{e_i, w_i\}$.

At inference (part (b) in Figure \ref{fig:word_confusion}), we wish to define the semantic similarity of a word in terms of the classifier's classes, which can be thought of as features.\footnote{Thus the choice of a different set of classes is a way of selecting different features  to describe the input word.} Now suppose we would like to compute the similarity of the new word ``burgundy'' to various colors. We extract the contextual embedding of  ``burgundy'' given the sentence \textit{``Burgundy is a deep reddish-brown shade inspired by wine''}, and  use the trained classifier to compute the probability that the \textit{``burgundy''}-embedding corresponds to each class $W=\{red,\ green,\ blue\}$. We then use the classifier's confusion matrix to understand which primary colors burgundy is similar to.  For example, the similarity of ``burgundy'' to  ``red'' is the probability our classifier assigns to the class \textit{``red''}.

More formally, the probability distribution predicted by the model, $\vec{p_j} \in \mathbb{R}^{|W|}$, is used to quantify the semantic similarity between $w_j$ (Burgundy) and each $w_i, \forall w_i \in W$ (in this case $W=$\{{red},\ green,\ blue\}).   Thus:
\begin{eqnarray}
\text{sim}_{\text{WC}}(w_i,w_j) \stackrel{\mathrm{def}}{=}
 p(w_i|e_j)
\end{eqnarray}
The set of distractor words chosen to train the initial classifier thus act as features that can be selected by the analyst to focus on a particular dimension or question.
 
Note that as with the example ``burgundy",  the input word at inference can be out-of-vocabulary with respect to the classifier, or the target word can be one of the classifier's classes (in which case we ignore the probability it assigns to that word and use the other $N-1$ features.)

\label{sec:cosine_vs_probe}
\begin{figure*}[h!]
    \centering
    \includegraphics[width=0.9\textwidth]{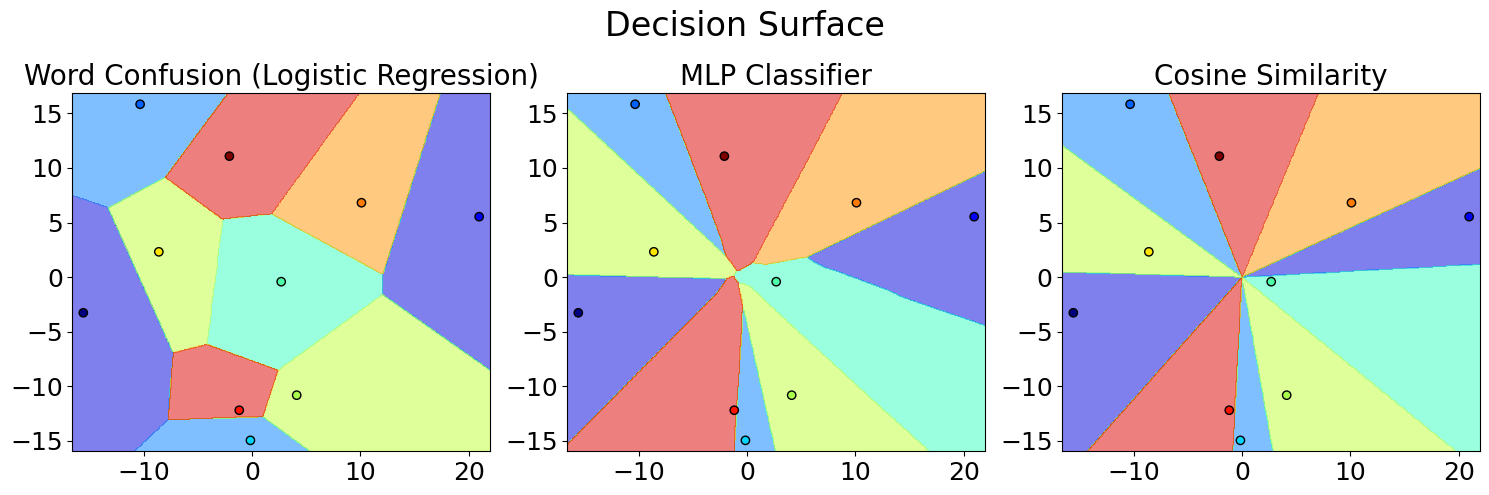}
    \caption{Differences in decision boundaries between \wc and cosine similarity. The $x$ and $y$ axes represent two dimensions of an artificially constructed set of data points. Note how cosine similarity's boundaries originate from the origin whereas {\wc}'s are not limited in the same way.}
    \label{fig_concept_prob_boundaries}
\end{figure*}

\subsection{Benchmarking \wc}
\label{sec:initial_eval}

The intuition behind \wc is that if it struggles to distinguish between contextual embeddings of \textit{burgundy} and \textit{red}, this could indicate they are similar. To test this hypothesis, we use \wc on three semantic similarity benchmarks. For each task, we trained a model using sentences from English Wikipedia. Our classes contained all the words from the benchmark. We then built word embeddings by averaging the last four hidden layers of BERT-base-cased (Details in Appendix~\ref{appendix:validation_details}). 

To calculate the similarity between two words $w_i, w_j$, we first extract all the sentences containing $w_i$ from English Wikipedia. We average the contextual token embeddings of $w_i$ using these sentences. This average token embedding was the input to the trained classifier (with classes containing all the words in the benchmark). We then use the probability \wc assigned to $w_i$  to set the similarity score between $w_i$ and $w_j$. We tested three benchmarks:

\begin{itemize}
\itemsep 0pt
    \item \textbf{MEN} contains 3000-word pairs annotated by 50 humans  based on their ``relatedness'' \cite{agirre-etal-2009-study}. For example \{berry, seed\}, \{game, hockey\}, and \{truck, vehicle\} received high relatedness scores, where \{hot, zombi\}, \{interior, mushroom\}, and \{bakery, zebra\} received low scores. 
To approximate human agreement, two annotators labeled all 3000 pairs on a 1-7 Likert scale; their Spearman correlation is 0.68, and the correlation of their average ratings with the general MEN scores is 0.84. 
%This high correlation suggests that MEN contains meaningful semantic ratings.
\item \textbf{WordSim353 (WS353)} contains 2000 word-pairs along with human-assigned association judgements \cite{bruni2014multimodal}. For example \{bank, money\}, \{Jerusalem, Israel\}, and \{Maradona, football\} received high scores whereas \{noon, string\}, \{sugar, approach\}, and \{professor, cucumber\} were ranked low. 
The authors report an inter-annotator agreement of 84\%. 

\item \textbf{SimLex} contains 1000 word-pairs and directly measures similarity, rather than relatedness or association \cite{hill-etal-2015-simlex}. The authors defined similarity as synonymy and instructed their annotators to rank accordingly. For example \{happy, glad\}, \{fee, payment\}, and \{wisdom, intelligence\} received high relatedness scores, where \{door, floor\}, \{trick, size\}, and \{old, new\} received low scores. 
Inter-rater agreement (the average of pairwise Spearman correlations between the ratings of all respondents) was reported as 0.67.

\end{itemize}

\begin{table}[]
    \centering
    \begin{tabular}{l|ccc} \toprule
         \diagbox[dir=NW]{{Method}}{{Dataset}} & {MEN} & {WS353} & {SimLex}\\ \midrule
         Cosine & {0.59} & 0.54 & 0.39 \\
         \makecell[l]{{\dedicationfont \textcolor{navyblue}{Word}}\\{{\dedicationfont \textcolor{navyblue}Confusion}}} & \textbf{0.66} &\textbf{0.67} & \textbf{0.44} \\ \bottomrule
    \end{tabular}
    \caption{Spearman's $\rho$ correlation between \wc and cosine similarity results as compared to humans. These three benchmarks focus on slightly different aspects of word similarity. We measure the correlation between human scores and cosine similarity between the language model embeddings versus {\wc}'s similarity scores. As can be seen, our method slightly outperforms cosine similarity.}
    \label{tab:semantic_similarity_results}
\end{table}

Across MEN, WS353, and SimLex, \wc slightly outperforms cosine similarity. This illustrates the meaningfulness of classification confusions, compared to cosine similarity. We note that our probability distribution spanned only the classes we chose in advance (all of the words in the dataset), which yields a different vocabulary compared to the original language model.

\section{Theoretical Intuitions}

In this section, we discuss the importance of word identifiability and how it enables the core mechanics of \wc, and discuss some theoretical differences between \wc and cosine similarity.  

\subsection{The Identifiability of Contextualized Word Embeddings}

\wc depends on the ability of a classifier to identify a word based on its contextual embedding; here we confirm that this classification task is indeed solvable, and examine some error cases to better understand it.

While contextualized word embeddings vary in their representation based on context, prior work showed that tokens of the same word still cluster together in geometric space \cite{zhou-etal-2022-problems}.

To test whether these boundaries are indeed learnable, we test how well a model can identify a contextualized word embedding after seeing one other example of the same word's contextualized embedding. We randomly sampled 26,000 words from English Wikipedia, trained 1000-class one-shot classifiers, and tested them on 10,000 examples (ten examples per class). Indeed, we found that the average test set accuracy on all our classifiers is 90\%, suggesting that the contextualized word embeddings are highly \textit{identifiable}. Thus, given an embedding, it is possible to identify its symbolic representation. See \ref{section:appendix_identity_probe} for experimental details.

\subsection{Differences Between \wc and Cosine Similarity}

\wc and cosine similarity give different kinds of distances.
We can see one way to visualize this in Figure~\ref{fig_concept_prob_boundaries}. Note the differences in the decision surface between {\wc} and cosine similarity: cosine boundaries emerge from the origin, whereas boundaries from \wc are not restricted in the same way.

Using a linear classifier in \wc also introduces new parameters that transform the input vectors into a different space, effectively redefining the notion of distance compared to the raw embeddings. To see this, consider two normalized 2-dimensional vectors, $x$ and $y$, and a real linear transformation, $A$ applied to each. Using the singular value decomposition (SVD) of $A= U \Sigma {V}^\intercal$, the singular values of $A$ (${\sigma}{u}{{v}^\intercal}$) allow us to rewrite the transformed vectors $Ax, Ay$ as ${\sigma_1}{u_1}{{v_1}^\intercal}{x_1} + {\sigma_2}{u_2}{{v_2}^\intercal}{x_2}$ and  ${\sigma_1}{u_1}{{v_1}^\intercal}{y_1} + {\sigma_2}{u_2}{{v_2}^\intercal}{y_2}$ respectively. 

The cosine distance between the transformed vectors is $1-({\sigma_1}^2({v_1}^{\intercal}x_1)({v_1}^{\intercal}y_1) + {\sigma_2}^2({v_2}^{\intercal}{x_2})({v_2}^{\intercal}{y_2}))$ compared to the original cosine distance $1 - (x_1y_1 + x_2y_2)$.\footnote{Terms cancel out as ${\sigma_1}{\sigma_1}=1$ and ${\sigma_1}{\sigma_2}=0$.} Similarly, the Euclidean distance between the transformed vectors is ${\sigma_1}{u_1v_1}^\intercal({x_1-y_1}) + {\sigma_2}{u_2v_2}^\intercal({x_2-y_2})$ compared to the original Euclidean distance of $(x_1-y_1)^2 + (x_2-y_2)^2$. In both cases, the distances between the two transformed vectors differ from the original vectors based on the linear transformation applied.

In other words, a linear transformation introduces additional parameters, allowing the model to reshape the geometry of word vectors and adjust the distances between words and their predicted semantic similarities.

Our method also shares some properties with cosine similarity.  Because linear classifiers learn a weight vector for each category that represents a kind of prototype of the category, the weight vectors learned by our classifier will be approximations of each word vector itself. Like cosine, our method thus computes similarity as the dot product between the word vector input and an errorful representation of the word vector encoded in the weights of the final classifier. What makes this approach effective is its reliance on small yet informative prediction errors that encode a meaningful signal, making the confusion matrix a source of linguistic insight.

% will geometry of the embeddings, giving a model additional parameters to represent targeted words and 
% we discuss differences between \wc and cosine similarity, arguing that feature-based similarity can produce more flexible decision boundaries, capture asymmetrical relations, highlight specific aspects of the analyzed word, and output more meaningful scores. 

\subsection{Advantageous Properties of \wc as a Similarity Measure}

Using a trainable linear classifier and analyzing its error signal for word-similarity purposes introduces a few advantages for measuring similarity:

\xhdr{Asymmetry} Human perceived similarity is not symmetric \cite{tversky1977features}. Yet cosine, like many distance functions commonly used to calculate semantic similarity, is symmetric. One of the advantages of using a model's confusion matrix for measuring semantic similarity is that these scores are \textit{asymmetric}; i.e., $p_{ij} \not= {p_{ji}}$. For example, \wc assigns lower probabilities for \textit{animal} being predicted as \textit{cat} than for \textit{cat} being predicted as \textit{animal}. The ability to measure asymmetric semantic similarity opens interesting new directions of understanding semantic similarity which are not possible with cosine.

\xhdr{Domain Adaptability} The fact that {\wc} requires training leads to more flexible similarity measures. Class selection enables measuring the semantic similarity of words relative to just a \textbf{subset} of features; we propose that this is particularly useful for practitioners who are interested in computing the similarity of words within a niche domain (we explore this in section \ref{sec:domain_spec}). 
% One limitation of our method is that it is self-supervised and would need to be retrained for out-of-domain words, which should still be quick given it is training a logistic regression 

\xhdr{Interpretability}
Probabilistic similarity measures have the advantage of being more interpretable for humans than non-probabilistic measures like cosine  \cite{sohangir2017improved}. Using a classifier's confusion matrix gives similarity scores that represent real probabilities. Moreover, since the choice of classifier classes is an implementation decision, one could choose them based on desired aspects of a word for a task. For example, we could interpret attitudes toward school by asking for the confusion matrix for the word ``school'' with a sentiment analysis classifier that contains the classes \{\textit{negative}, \textit{positive}\}, or the classes \{\textit{fun}, \textit{work}\}. 
\section{Real-World Data}
\label{sec:domain_spec}

\wc is a new similarity measuring tool that could assist in understanding real-world data and trends. In this section, we focus on two applications of {\wc} -- its ability to serve as a feature extractor and to detect temporal trends in word meaning.   

\subsection{\wc for Feature Classification}
\label{sec:validation_experiments}

\begin{table*}[h!]
    \centering
    \resizebox{\textwidth}{!}{%
    \begin{tabular}{lccccc} \toprule
    \textbf{Experiment} & \textbf{\wc} & \textbf{Cosine 1} & \textbf{Cosine 2} & \textbf{Cosine 3} & \textbf{Ave. Cosine} \\ \midrule
     Sentiment Classification & \textbf{0.79} & 0.75 & 0.71 & 0.84 & 0.73 \\
     Grammatical Gender (Italian) & \textbf{0.93} & 0.80 & 0.80 & 0.71 & 0.77 \\
     Grammatical Gender (French) & 0.85 & \textbf{0.86} & \textbf{0.86} & 0.83 & 0.85 \\
     ConceptNet Domain (Fashion-Gaming) & 0.90 & \textbf{0.93} & \textbf{0.93} & 0.90 & 0.92 \\
     ConceptNet Domain (Sea-Land Animals) & \textbf{0.83} & 0.79 & 0.80 & 0.61 & 0.73 \\
     \midrule
     Average & \textbf{0.86} & 0.83 & 0.82 & 0.76 & 0.80 \\ \bottomrule
    \end{tabular}
    }
    \caption{Macro-F1 for \wc and cosine similarity across a variety of feature classification tasks. We operationalize cosine similarity in three ways: 1) the distance between the centroids of the seed words and the target words 2) the average distance each of the target word to the centroid of the seed words 3) the average distance of each target word to each seed word (no centroids).}
    \label{tab:results}
\end{table*}

\wc can be used to define out-of-domain word classes, i.e. when $w_j \not\in W$. Using our earlier example, if the classes of \wc are the features \{\textit{positive}, \textit{negative}\}, given an out-of-domain word like \textit{school}, we can use the confusion matrix to represent the embedding for \textit{school} as a mixture of the classes the model is familiar with, i.e., \{\textit{positive}, \textit{negative}\}.

Following this intuition, we test whether \wc can use features as classes to identify objects' membership to these classes accurately. We used the following tasks: 

\noindent \textbf{Sentiment classification} using the NRC corpus \cite{pang-etal-2002-thumbs, mohammad-etal-2013-nrc}. The goal is to classify words according to their sentiment (either positive or negative). The words were manually annotated based on their emotional association (e.g., ``trophy'' is positive, whereas ``flu'' is negative).

\noindent \textbf{Grammatical gender} classification of nouns \cite{sahai-sharma-2021-predicting}. We tested \wc using two languages -- Italian and French. The goal is to classify words according to their grammatical gender per language. For example, ``flower'' is feminine in French and masculine in Italian. 

\noindent \textbf{Domain classification} using ConceptNet categories \cite{dalvi2022discovering}. The goal is to classify words to their correct ConceptNet class. We used two domain pairs: Fashion-Gaming is about classifying whether a word belongs to the fashion domain or the gaming domain; in Sea-Land, the goal is to predict if an animal is a sea or land animal.

For each task, we hand-select meaningful words as classes for the classifier and use terms from the lexicon as test embeddings. For example, for sentiment classification we first use the seed words \textit{positive} and \textit{negative} as our classes and collect occurrences from a corpus, extract the embeddings train the concept prober to recognize \textit{positive} and \textit{negative}. Finally, we then use \wc to classify all the terms in the NRC lexicon (our target words). We define the label using the class with the highest probability for the word. Details of each experiment are available in in the Appendix \ref{sec:seeds}.

Across all three tasks, we find that \wc is successful in feature-based classification using a few seed word training examples. Compared to cosine similarity, we achieve a macro-F1 of 86\% compared to 80\% (see table \ref{tab:results}). 

\xhdr{Embedding Meaning vs. Properties}
It is important to distinguish between embedding a word for its overall meaning (e.g., whether it conveys a positive or negative sentiment) versus embedding it to capture a specific property (e.g., gender, formality). While \wc supports both, this distinction is crucial when interpreting the results and determining the appropriate transformations for different tasks.

\subsection{What Is A Revolution?}

We now examine whether \wc could be used as a tool for studying concepts in a way that aids humanistic or social science investigations. By collaborating with the fourth author, a scholar of French literature and history, we investigate historical changes in the meaning of the French word \textit{``révolution''}. Together, we used \wc to test a prominent hypothesis of how the meaning of the word and concept of revolution changed \cite{baker_1990}: that the meaning of \textit{``révolution''} in the early years of the French Revolution was more associated with \textit{popular} action, but later become identified with \textit{state} actions.

We constructed a set of French words associated with the people (\{\textit{peuple}, \textit{populaire}, ...\}) and the state (\{\textit{conseil}, \textit{gouvernement}, ...\}).\footnote{Note on choice of seed words: we are tracking changes in the meaning of \textit{``révolution''} between 1789 and 1793 thus only looking at the vocabulary used during the French Revolution. Although the connection between \textit{``peuple''} and \textit{``révolution''} could be found before July 14, 1789, it is in the aftermath of that date that this connection became the primary one. Prior to this, the delegates of the National Assembly in Versailles had claimed they had been leading the \textit{``révolution''}.} These seed words were used as classes for our classifier, which we trained on different temporal segments to observe the temporal change in our concept of interest. Our corpora are the \textit{Archives Parlementaires}, transcripts of parliamentary speeches during a time that contains moments of both emancipation and elite control of political processes.\footnote{\url{https://sul-philologic.stanford.edu/philologic/archparl/}} The corpus contains 9,628 speeches and 54,460,150 words from the years 1789-1793. Within this corpus, the term \textit{``révolution''} appears 2,206 times across 218 speeches, with a contextual basis of 90,138 words.

\begin{figure}[h!]
    \includegraphics[width=.45\textwidth]{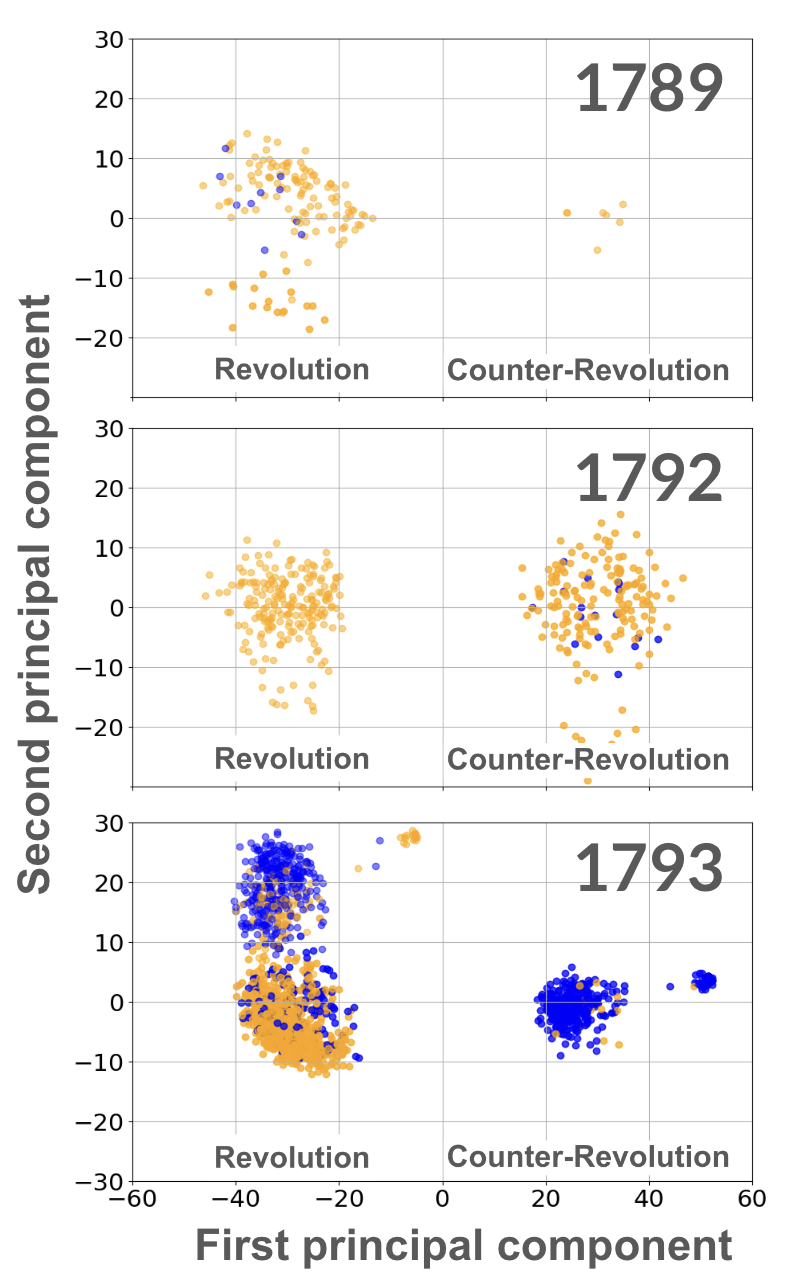}
    \caption{In 1789, the word \textit{``revolution''} was primarily associated with popular action (represented in orange). In 1792 \textit{``revolution''} was now also seen as something that the government should lead (represented in blue) found in the \textit{``counter-revolution''} cluster. In 1793, this new governmental meaning had spread back to the word \textit{``revolution''} itself.
    % \cnote{TODO: improve visualization. Some text is not readable. Add legend}
    }
    \label{fig:revolution}
\end{figure}

We color-code the classes orange as \textit{``peuple'' } (\textit{``the people''}) and blue as \textit{``gouvernement''} (\textit{``the state''}) and project the embeddings down to a 2-dimensional space and visualize the results (Figure \ref{fig:revolution}).

We find that, in 1789, the word \textit{``révolution''} was primarily associated with popular action, the most famous example of which was the storming of the Bastille. In 1792, another definition became common: \textit{``révolution''} was now also seen as something that the government should lead. 

Interestingly, the first use of \textit{``révolution''} to be associated with governmental action is in fact around the term \textit{``counter-revolution''}. The word embeddings of \textit{``counter-révolution''} are predicted to be associated with the state, indicating that it was primarily when talking about threats to, and enemies of, the revolution, that politicians suggested transferring more power to the state. Jumping forward to 1793, this new governmental meaning is predicted for the word embeddings of \textit{``révolution''} itself. Our findings suggest that the goal of repressing counter-revolutionaries is what associated the term \textit{``révolution''} with governmental action. In other words, once revolutionaries became more concerned about tracking down their enemies, they granted to the government the same kind of extra-legal power that had originally only been the prerogative of the people in arms.\footnote{While the proclamation of the republic and the introduction of the new calendar are related to the idea of revolution, the conceptual shifts that we've identified appears prior to both of these events. The revolutionary calendar was not introduced until October 1793, meanwhile declaration of the Republic occurred in September 1792. The emergence of \textit{``counterrevolution''} as a state problem predates these events, confirming that neither play a role in introducing the newer understanding of ``revolution'' as a state-driven process.}

Our findings are consistent with historians' hypothesis that the meaning of revolution in the early years of the French Revolution is most closely aligned with the concept of the people and this gradually shifts as the revolution continues \cite{sewell2005logics,edelstein2012we}. Furthermore, our model allows us to uncover a potential causal story for this shift in the meaning; that the state sense of revolution first actually started with counter-revolution. This is a novel discovery in our understanding of the French Revolution; future humanistic work should use other methods to confirm this proposed causal link to counter-revolutionaries.

 See  Appendix~\ref{sec:finance_experiment} for another, more preliminary  social scientific application of \wc, in this case to study financial trends.  
\section{Related Work on Cultural Change}
\label{sec:related_work}

Both static and contextualized embedding spaces contain semantically meaning dimensions that align with high-level linguistic and cultural features \citep{bolukbasi2016man, DBLP:journals/corr/abs-1906-02715}. These embeddings have enabled a large number of quantitative analyses of temporal shifts in meaning and links to cultural or social scientific variables. For example early on, using static embeddings, \citet{hamilton2016cultural} measured linguistic drifts in global semantic space as well as cultural shifts in particular local semantic neighborhoods. \citet{garg2018word} demonstrated that changes in word embeddings correlated with demographic and occupation shifts through the 1900s.

Analyzes of contextualized embeddings have identified semantic axes based on pairs of ``seed words'' or ``poles'' \citep{soler2020bert, lucy2022discovering, grand2022semantic}. Across the temporal dimension, such axes can measure the evolution of gender and class \citep{kozlowski2019geometry}, internet slang \citep{keidar-etal-2022-slangvolution}, and more \citep{madani2023measuring, lyu2023representation, erk2024adjusting}. \citet{bravzinskas2017embedding} proposes a probabilistic measure for lexical similarity. 

It's also instructive to consider the similarity of our method  with tasks like word sense disambiguation (WSD) and named entity recognition (NER). The central idea behind \wc of mapping from embeddings to categories are also found in NER and WSD. What differs is the dynamic nature of the categories. Where NER focuses on pre-defined concept hierarchies and WSD on pre-defined senses per word,  \wc  focuses on a coherent but dynamic grouping of words that is interpretable for a given task.

% an important direction for future work in computational social science (see 3.1.10 in \citet{ziems2023can}).

% \subsection{Semantic change}

% Survey \citep{de2024survey} (should prob look more)

% Diachronic word embeddings 

%  \citet{di2019training}

% Similarly to named entity recognition, \ac{ourmethod} attempts to map words to classes that \textit{might be} types (such as ORGANIZATION or PLACE). However, \ac{ourmethod} is partially self-supervised and it is entirely user-driven.

% Similarly to semantic change detection, \ac{ourmethod} attempts to capture the usage of a word in different contexts. However, \ac{ourmethod} offers the possibility of defining the axis (the seed words) onto which the user wants to project words. 

\section{Discussion and Conclusion}
% In this paper, we reframe the task of semantic similarity from one of measuring distances to one of classification confusions. Our method is self-supervised and allows for researchers and downstream practitioners to measure the similarity of words with ease. We illustrate the performance of the identity probe on a number of word similarity tasks as well as provide examples of other tasks the identity probe can generalize to. In addition to the identity probe's high performance, we also find that the identity probe is able to produce asymmetric measures of similarity, opening a new line of future research. Lastly, we conclude with theoretical intuition of how a simple reframing of the problem results in new decision boundaries of similarity. We're excited for the NLP and broader community to use the identity probe as a measure of semantic similarity and discover other new ways in which the probe can be used for classification tasks.

In this paper, we reframe the task of semantic similarity from one of measuring distances to one of classification confusion. This formulation highlights the context-dependency of similarity judgments, meanwhile avoiding the pitfalls of geometric similarity measures \cite{evers2014revisiting}.

This new framing of semantic similarity in terms of classification confusion introduces new properties that are inspired by cognitive models of similarity \cite{tversky1977features} and accounts for the asymmetric nature of semantic similarity, captures different aspects of both similarity and multi-faceted words and ofter a measure that has interpretability benefits. 

Our proof-of-concept method, \wc, demonstrates the practical applicability and effectiveness of this reframing. Empirical results show that it outperforms cosine similarity on standard datasets.
For computational social science or cultural analytics applications, \wc can serve as a way to learn to represent words using target features (e.g., ``school'' in terms of \{\textit{positive}, \textit{negative}\}, and can be used to trace the meaning of a word as a function of time (like the word ``r\'{e}volution'').

The theoretical underpinnings of \wc allow it to learn complex word identity boundaries and capture the directional nature of similarity, offering a richer and more flexible framework for understanding word meanings. 

%To conclude, we reframe semantic similarity using classification confusion, to align it better with psychology literature. We implement a proof-of-concept framework that highlights the desired aspect(s) of a multi-faceted word by classifying it based on better-defined adjacent terms. 

While we implemented \wc as a linear classifier, the method naturally extends to capturing non-linear relationships among embedding components by replacing the linear projection with neural networks. Investigating whether the error function preserves its useful properties in non-linear settings remains an open question for future work.

While our experiments are preliminary and the space of possible similarity measures is enormous, we hope this reimagining of semantic similarity will inspire the development of new tools that better capture the multi-faceted and dynamic nature of language, advancing the fields of computational social science and cultural analytics and beyond.

\section*{Limitations}
Our proof-of-concept suggests a promising path where cosine similarity can be replaced by a more sophisticated method that involves self-supervision. However, the boost in performance also comes with some caveats. Because \wc is a supervised classifier, it requires an extra training step that simple cosine doesn't require.  Furthermore, potential users will need a basic understanding of model training and the pitfalls of over-fitting data.

As mentioned earlier, while we implemented \wc as a linear classifier, the method naturally extends to non-linear models. Additionally, various transformations commonly applied to embeddings before measuring distances \cite{Mu2018AllbuttheTopSA} can also be incorporated prior to using \wc, as our method relies on the resulting error signal to assess word similarity. Although non-linear models offer a promising direction, we have not yet examined whether the error function preserves its useful properties in such settings—an important avenue for future work. Introducing non-linearity into the classifier is known to alter its behavior in various ways, but its impact on confusion-based similarity remains uncertain. Further research is needed to evaluate its potential advantages and limitations.

Another key limitation of our approach is that we used three simple implementations of cosine similarity without exploring many possible augmentations to cosine, like normalizing it across the dataset (as was shown to be effective by \cite{timkey2021all}). Further refining both our classifier and cosine similarity implementations could lead to improved results for both,  as well as a deeper understanding of \wc.

Another important limitation of our analysis is that our results might be affected by the choice of seed words and the mechanisms on how we sample the ones used to represent the concepts. Changing seed words can impact the similarities. While we explored different sets of seed words without seeing drastic changes in results, a robust evaluation of the effect of different seed words should be considered in future work.

Lastly, we are not aware if changing the model used to create the embeddings can degrade the performance.

\section*{Ethics Statement}
As with all language technologies, there are a number of ethical concerns surrounding their usage and societal impact. It is likely that with this method, the biases known in contextualized embeddings can continue to propagate through downstream tasks, leading to representation or allocation harms. Additionally, the use of large language models for building contextualized embeddings is expensive and requires time and energy resources. To our knowledge, the method we have developed does not exacerbate any of these pre-existing ethical concerns but we recognize our work here also does not mitigate or avoid them.

\section*{Acknowledgments}
We thank Dallas Card, Nelson Liu, Kyle Hsu, Amelia Hardy, Kawin Ethayarajh, and Tianyi Zhang for their helpful feedback and discussion. This work was supported in part by the NSF via award number IIS-2128145, by the Hoffman–Yee Research Grants Program and the Stanford Institute for Human-Centered Artificial Intelligence, and by the Koret Foundation grant for Smart Cities and Digital Living.

\bibliography{anthology,custom}

\begin{thebibliography}{50}
\expandafter\ifx\csname natexlab\endcsname\relax\def\natexlab#1{#1}\fi

\bibitem[{Agirre et~al.(2009)Agirre, Alfonseca, Hall, Kravalova, Pa{\c{s}}ca, and Soroa}]{agirre-etal-2009-study}
Eneko Agirre, Enrique Alfonseca, Keith Hall, Jana Kravalova, Marius Pa{\c{s}}ca, and Aitor Soroa. 2009.
\newblock \href {https://aclanthology.org/N09-1003} {A study on similarity and relatedness using distributional and {W}ord{N}et-based approaches}.
\newblock In \emph{Proceedings of Human Language Technologies: The 2009 Annual Conference of the North {A}merican Chapter of the Association for Computational Linguistics}, pages 19--27, Boulder, Colorado. Association for Computational Linguistics.

\bibitem[{Baker(1990)}]{baker_1990}
Keith~Michael Baker. 1990.
\newblock \href {https://doi.org/10.1017/CBO9780511625527} {\emph{Inventing the French Revolution: Essays on French Political Culture in the Eighteenth Century}}.
\newblock Ideas in Context. Cambridge University Press.

\bibitem[{Bhattacharya et~al.(2020)Bhattacharya, Ghosh, Pal, and Ghosh}]{bhattacharya2020methods}
Paheli Bhattacharya, Kripabandhu Ghosh, Arindam Pal, and Saptarshi Ghosh. 2020.
\newblock Methods for computing legal document similarity: A comparative study.
\newblock \emph{arXiv preprint arXiv:2004.12307}.

\bibitem[{Blevins and Zettlemoyer(2020)}]{blevins-zettlemoyer-2020-moving}
Terra Blevins and Luke Zettlemoyer. 2020.
\newblock \href {https://doi.org/10.18653/v1/2020.acl-main.95} {Moving down the long tail of word sense disambiguation with gloss informed bi-encoders}.
\newblock In \emph{Proceedings of the 58th Annual Meeting of the Association for Computational Linguistics}, pages 1006--1017, Online. Association for Computational Linguistics.

\bibitem[{Bolukbasi et~al.(2016)Bolukbasi, Chang, Zou, Saligrama, and Kalai}]{bolukbasi2016man}
Tolga Bolukbasi, Kai-Wei Chang, James~Y Zou, Venkatesh Saligrama, and Adam~T Kalai. 2016.
\newblock Man is to computer programmer as woman is to homemaker? debiasing word embeddings.
\newblock \emph{Advances in neural information processing systems}, 29.

\bibitem[{Bra{\v{z}}inskas et~al.(2017)Bra{\v{z}}inskas, Havrylov, and Titov}]{bravzinskas2017embedding}
Arthur Bra{\v{z}}inskas, Serhii Havrylov, and Ivan Titov. 2017.
\newblock Embedding words as distributions with a bayesian skip-gram model.
\newblock \emph{arXiv preprint arXiv:1711.11027}.

\bibitem[{Bruni et~al.(2014)Bruni, Tran, and Baroni}]{bruni2014multimodal}
Elia Bruni, Nam-Khanh Tran, and Marco Baroni. 2014.
\newblock Multimodal distributional semantics.
\newblock \emph{Journal of artificial intelligence research}, 49:1--47.

\bibitem[{Caliskan et~al.(2017)Caliskan, Bryson, and Narayanan}]{caliskan2017semantics}
Aylin Caliskan, Joanna~J Bryson, and Arvind Narayanan. 2017.
\newblock Semantics derived automatically from language corpora contain human-like biases.
\newblock \emph{Science}, 356(6334):183--186.

\bibitem[{{Center for Bibliographic Studies and Research, University of California, Riverside}(2024)}]{CDNC2024data}
{Center for Bibliographic Studies and Research, University of California, Riverside}. 2024.
\newblock Courtesy of the california digital newspaper collection.
\newblock Data retrieved from World Development Indicators, \url{http://cdnc.ucr.edu}.

\bibitem[{Coenen et~al.(2019)Coenen, Reif, Yuan, Kim, Pearce, Vi{\'{e}}gas, and Wattenberg}]{DBLP:journals/corr/abs-1906-02715}
Andy Coenen, Emily Reif, Ann Yuan, Been Kim, Adam Pearce, Fernanda~B. Vi{\'{e}}gas, and Martin Wattenberg. 2019.
\newblock \href {http://arxiv.org/abs/1906.02715} {Visualizing and measuring the geometry of {BERT}}.
\newblock \emph{CoRR}, abs/1906.02715.

\bibitem[{Dalvi et~al.(2022)Dalvi, Khan, Alam, Durrani, Xu, and Sajjad}]{dalvi2022discovering}
Fahim Dalvi, Abdul~Rafae Khan, Firoj Alam, Nadir Durrani, Jia Xu, and Hassan Sajjad. 2022.
\newblock \href {https://openreview.net/forum?id=POTMtpYI1xH} {Discovering latent concepts learned in {BERT}}.
\newblock In \emph{International Conference on Learning Representations}.

\bibitem[{Edelstein(2012)}]{edelstein2012we}
Dan Edelstein. 2012.
\newblock Do we want a revolution without revolution? reflections on political authority.
\newblock \emph{French Historical Studies}, 35(2):269--289.

\bibitem[{Erk and Apidianaki(2024)}]{erk2024adjusting}
Katrin Erk and Marianna Apidianaki. 2024.
\newblock Adjusting interpretable dimensions in embedding space with human judgments.
\newblock \emph{arXiv preprint arXiv:2404.02619}.

\bibitem[{Ethayarajh(2019)}]{ethayarajh2019contextual}
Kawin Ethayarajh. 2019.
\newblock How contextual are contextualized word representations? comparing the geometry of bert, elmo, and gpt-2 embeddings.
\newblock \emph{arXiv preprint arXiv:1909.00512}.

\bibitem[{Ettinger and Linzen(2016)}]{ettinger2016evaluating}
Allyson Ettinger and Tal Linzen. 2016.
\newblock Evaluating vector space models using human semantic priming results.
\newblock In \emph{Proceedings of the 1st workshop on evaluating vector-space representations for NLP}, pages 72--77.

\bibitem[{Evers and Lakens(2014)}]{evers2014revisiting}
Ellen~RK Evers and Dani{\"e}l Lakens. 2014.
\newblock Revisiting tversky's diagnosticity principle.
\newblock \emph{Frontiers in Psychology}, 5:57776.

\bibitem[{Garg et~al.(2018)Garg, Schiebinger, Jurafsky, and Zou}]{garg2018word}
Nikhil Garg, Londa Schiebinger, Dan Jurafsky, and James Zou. 2018.
\newblock Word embeddings quantify 100 years of gender and ethnic stereotypes.
\newblock \emph{Proceedings of the National Academy of Sciences}, 115(16):E3635--E3644.

\bibitem[{Grand et~al.(2022)Grand, Blank, Pereira, and Fedorenko}]{grand2022semantic}
Gabriel Grand, Idan~Asher Blank, Francisco Pereira, and Evelina Fedorenko. 2022.
\newblock Semantic projection recovers rich human knowledge of multiple object features from word embeddings.
\newblock \emph{Nature human behaviour}, 6(7):975--987.

\bibitem[{Hamilton et~al.(2016)Hamilton, Leskovec, and Jurafsky}]{hamilton2016cultural}
William~L Hamilton, Jure Leskovec, and Dan Jurafsky. 2016.
\newblock Cultural shift or linguistic drift? comparing two computational measures of semantic change.
\newblock In \emph{Proceedings of the conference on empirical methods in natural language processing. Conference on empirical methods in natural language processing}, volume 2016, page 2116. NIH Public Access.

\bibitem[{Hewitt and Liang(2019)}]{hewitt-liang-2019-designing}
John Hewitt and Percy Liang. 2019.
\newblock \href {https://doi.org/10.18653/v1/D19-1275} {Designing and interpreting probes with control tasks}.
\newblock In \emph{Proceedings of the 2019 Conference on Empirical Methods in Natural Language Processing and the 9th International Joint Conference on Natural Language Processing (EMNLP-IJCNLP)}, pages 2733--2743, Hong Kong, China. Association for Computational Linguistics.

\bibitem[{Hill et~al.(2015)Hill, Reichart, and Korhonen}]{hill-etal-2015-simlex}
Felix Hill, Roi Reichart, and Anna Korhonen. 2015.
\newblock \href {https://doi.org/10.1162/COLI_a_00237} {{S}im{L}ex-999: Evaluating semantic models with (genuine) similarity estimation}.
\newblock \emph{Computational Linguistics}, 41(4):665--695.

\bibitem[{Johri et~al.(2011)Johri, Ramage, McFarland, and Jurafsky}]{johri-etal-2011-study}
Nikhil Johri, Daniel Ramage, Daniel McFarland, and Daniel Jurafsky. 2011.
\newblock \href {https://aclanthology.org/W11-1516} {A study of academic collaborations in computational linguistics using a latent mixture of authors model}.
\newblock In \emph{Proceedings of the 5th {ACL}-{HLT} Workshop on Language Technology for Cultural Heritage, Social Sciences, and Humanities}, pages 124--132, Portland, OR, USA. Association for Computational Linguistics.

\bibitem[{Keidar et~al.(2022)Keidar, Opedal, Jin, and Sachan}]{keidar-etal-2022-slangvolution}
Daphna Keidar, Andreas Opedal, Zhijing Jin, and Mrinmaya Sachan. 2022.
\newblock \href {https://doi.org/10.18653/v1/2022.acl-long.101} {Slangvolution: {A} causal analysis of semantic change and frequency dynamics in slang}.
\newblock In \emph{Proceedings of the 60th Annual Meeting of the Association for Computational Linguistics (Volume 1: Long Papers)}, pages 1422--1442, Dublin, Ireland. Association for Computational Linguistics.

\bibitem[{Kozlowski et~al.(2019)Kozlowski, Taddy, and Evans}]{kozlowski2019geometry}
Austin~C Kozlowski, Matt Taddy, and James~A Evans. 2019.
\newblock The geometry of culture: Analyzing the meanings of class through word embeddings.
\newblock \emph{American Sociological Review}, 84(5):905--949.

\bibitem[{Liu et~al.(2019)Liu, Gardner, Belinkov, Peters, and Smith}]{liu-etal-2019-linguistic}
Nelson~F. Liu, Matt Gardner, Yonatan Belinkov, Matthew~E. Peters, and Noah~A. Smith. 2019.
\newblock \href {https://doi.org/10.18653/v1/N19-1112} {Linguistic knowledge and transferability of contextual representations}.
\newblock In \emph{Proceedings of the 2019 Conference of the North {A}merican Chapter of the Association for Computational Linguistics: Human Language Technologies, Volume 1 (Long and Short Papers)}, pages 1073--1094, Minneapolis, Minnesota. Association for Computational Linguistics.

\bibitem[{Lucy et~al.(2022)Lucy, Tadimeti, and Bamman}]{lucy2022discovering}
Li~Lucy, Divya Tadimeti, and David Bamman. 2022.
\newblock \href {https://doi.org/10.18653/v1/2022.emnlp-main.228} {Discovering differences in the representation of people using contextualized semantic axes}.
\newblock In \emph{Proceedings of the 2022 Conference on Empirical Methods in Natural Language Processing}, pages 3477--3494.

\bibitem[{Lyu et~al.(2023)Lyu, Apidianaki, and Callison-burch}]{lyu2023representation}
Qing Lyu, Marianna Apidianaki, and Chris Callison-burch. 2023.
\newblock \href {https://doi.org/10.18653/v1/2023.starsem-1.32} {Representation of lexical stylistic features in language models{'} embedding space}.
\newblock In \emph{Proceedings of the 12th Joint Conference on Lexical and Computational Semantics (*SEM 2023)}, pages 370--387.

\bibitem[{Madani et~al.(2023)Madani, Bandyopadhyay, Swire-Thompson, Yoder, and Joseph}]{madani2023measuring}
Navid Madani, Rabiraj Bandyopadhyay, Briony Swire-Thompson, Michael~Miller Yoder, and Kenneth Joseph. 2023.
\newblock \href {http://arxiv.org/abs/2305.09548} {Measuring social dimensions of self-presentation in social media biographies with an identity-based approach}.

\bibitem[{Manzini et~al.(2019)Manzini, Yao~Chong, Black, and Tsvetkov}]{manzini-etal-2019-black}
Thomas Manzini, Lim Yao~Chong, Alan~W Black, and Yulia Tsvetkov. 2019.
\newblock \href {https://doi.org/10.18653/v1/N19-1062} {Black is to criminal as caucasian is to police: Detecting and removing multiclass bias in word embeddings}.
\newblock In \emph{Proceedings of the 2019 Conference of the North {A}merican Chapter of the Association for Computational Linguistics: Human Language Technologies, Volume 1 (Long and Short Papers)}, pages 615--621, Minneapolis, Minnesota. Association for Computational Linguistics.

\bibitem[{Martinc et~al.(2020)Martinc, Kralj~Novak, and Pollak}]{martinc-etal-2020-leveraging}
Matej Martinc, Petra Kralj~Novak, and Senja Pollak. 2020.
\newblock \href {https://aclanthology.org/2020.lrec-1.592} {Leveraging contextual embeddings for detecting diachronic semantic shift}.
\newblock In \emph{Proceedings of the Twelfth Language Resources and Evaluation Conference}, pages 4811--4819, Marseille, France. European Language Resources Association.

\bibitem[{Mohammad et~al.(2013)Mohammad, Kiritchenko, and Zhu}]{mohammad-etal-2013-nrc}
Saif Mohammad, Svetlana Kiritchenko, and Xiaodan Zhu. 2013.
\newblock \href {https://aclanthology.org/S13-2053} {{NRC}-{C}anada: Building the state-of-the-art in sentiment analysis of tweets}.
\newblock In \emph{Second Joint Conference on Lexical and Computational Semantics (*{SEM}), Volume 2: Proceedings of the Seventh International Workshop on Semantic Evaluation ({S}em{E}val 2013)}, pages 321--327, Atlanta, Georgia, USA. Association for Computational Linguistics.

\bibitem[{Mu et~al.(2018)Mu, Bhat, and Viswanath}]{Mu2018AllbuttheTopSA}
Jiaqi Mu, S.~Bhat, and Pramod Viswanath. 2018.
\newblock All-but-the-top: Simple and effective postprocessing for word representations.
\newblock \emph{ICLR}, abs/1702.01417.

\bibitem[{Nematzadeh et~al.(2017)Nematzadeh, Meylan, and Griffiths}]{nematzadeh2017evaluating}
Aida Nematzadeh, Stephan~C Meylan, and Thomas~L Griffiths. 2017.
\newblock Evaluating vector-space models of word representation, or, the unreasonable effectiveness of counting words near other words.
\newblock In \emph{CogSci}.

\bibitem[{New et~al.(2004)New, Pallier, Brysbaert, and Ferrand}]{npbf04}
B.~New, C.~Pallier, M.~Brysbaert, and L.~Ferrand. 2004.
\newblock \href {http://arxiv.org/abs/http://www.lexique.org/?page_id=294} {Lexique 2 : A new french lexical database}.
\newblock \emph{Behavior Research Methods, Instruments, \& Computers}, 36(3):516--524.

\bibitem[{Pang et~al.(2002)Pang, Lee, and Vaithyanathan}]{pang-etal-2002-thumbs}
Bo~Pang, Lillian Lee, and Shivakumar Vaithyanathan. 2002.
\newblock \href {https://doi.org/10.3115/1118693.1118704} {Thumbs up? sentiment classification using machine learning techniques}.
\newblock In \emph{Proceedings of the 2002 Conference on Empirical Methods in Natural Language Processing ({EMNLP} 2002)}, pages 79--86. Association for Computational Linguistics.

\bibitem[{Pescuma et~al.(2021)Pescuma, Zanini, Crepaldi, and Franzon}]{pescuma2021form}
Valentina~Nicole Pescuma, Chiara Zanini, Davide Crepaldi, and Francesca Franzon. 2021.
\newblock Form and function: A study on the distribution of the inflectional endings in italian nouns and adjectives.
\newblock \emph{Frontiers in Psychology}, page 4422.

\bibitem[{Pilehvar and Camacho-Collados(2019)}]{pilehvar-camacho-collados-2019-wic}
Mohammad~Taher Pilehvar and Jose Camacho-Collados. 2019.
\newblock \href {https://doi.org/10.18653/v1/N19-1128} {{W}i{C}: the word-in-context dataset for evaluating context-sensitive meaning representations}.
\newblock In \emph{Proceedings of the 2019 Conference of the North {A}merican Chapter of the Association for Computational Linguistics: Human Language Technologies, Volume 1 (Long and Short Papers)}, pages 1267--1273, Minneapolis, Minnesota. Association for Computational Linguistics.

\bibitem[{R{\'\i}os et~al.(2012)R{\'\i}os, Silva, and Aguilera}]{rios2012dissimilarity}
Sebasti{\'a}n~A R{\'\i}os, Roberto~A Silva, and Felipe Aguilera. 2012.
\newblock A dissimilarity measure for automate moderation in online social networks.
\newblock In \emph{Proceedings of the 4th International Workshop on Web Intelligence \& Communities}, pages 1--9.

\bibitem[{Sahai and Sharma(2021)}]{sahai-sharma-2021-predicting}
Saumya Sahai and Dravyansh Sharma. 2021.
\newblock \href {https://doi.org/10.18653/v1/2021.sigtyp-1.9} {Predicting and explaining {F}rench grammatical gender}.
\newblock In \emph{Proceedings of the Third Workshop on Computational Typology and Multilingual NLP}, pages 90--96, Online. Association for Computational Linguistics.

\bibitem[{Sewell~Jr. and Sewell(2005)}]{sewell2005logics}
William~H. Sewell~Jr. and William~Hamilton Sewell. 2005.
\newblock \emph{Logics of history: Social theory and social transformation}.
\newblock University of Chicago Press.

\bibitem[{Sitikhu et~al.(2019)Sitikhu, Pahi, Thapa, and Shakya}]{sitikhu2019comparison}
Pinky Sitikhu, Kritish Pahi, Pujan Thapa, and Subarna Shakya. 2019.
\newblock A comparison of semantic similarity methods for maximum human interpretability.
\newblock In \emph{2019 artificial intelligence for transforming business and society (AITB)}, volume~1, pages 1--4. IEEE.

\bibitem[{Sohangir and Wang(2017)}]{sohangir2017improved}
Sahar Sohangir and Dingding Wang. 2017.
\newblock Improved sqrt-cosine similarity measurement.
\newblock \emph{Journal of Big Data}, 4:1--13.

\bibitem[{Soler and Apidianaki(2020)}]{soler2020bert}
Aina~Garí Soler and Marianna Apidianaki. 2020.
\newblock \href {http://arxiv.org/abs/2010.02686} {Bert knows punta cana is not just beautiful, it's gorgeous: Ranking scalar adjectives with contextualised representations}.

\bibitem[{Timkey and Van~Schijndel(2021)}]{timkey2021all}
William Timkey and Marten Van~Schijndel. 2021.
\newblock All bark and no bite: Rogue dimensions in transformer language models obscure representational quality.
\newblock \emph{arXiv preprint arXiv:2109.04404}.

\bibitem[{Tversky(1977)}]{tversky1977features}
Amos Tversky. 1977.
\newblock Features of similarity.
\newblock \emph{Psychological review}, 84(4):327.

\bibitem[{Vilnis and McCallum(2014)}]{vilnis2014word}
Luke Vilnis and Andrew McCallum. 2014.
\newblock Word representations via gaussian embedding.
\newblock \emph{arXiv preprint arXiv:1412.6623}.

\bibitem[{Zhang and Bowman(2018)}]{zhang-bowman-2018-language}
Kelly Zhang and Samuel Bowman. 2018.
\newblock \href {https://doi.org/10.18653/v1/W18-5448} {Language modeling teaches you more than translation does: Lessons learned through auxiliary syntactic task analysis}.
\newblock In \emph{Proceedings of the 2018 {EMNLP} Workshop {B}lackbox{NLP}: Analyzing and Interpreting Neural Networks for {NLP}}, pages 359--361, Brussels, Belgium. Association for Computational Linguistics.

\bibitem[{Zhou et~al.(2022{\natexlab{a}})Zhou, Ethayarajh, Card, and Jurafsky}]{zhou2022problems}
Kaitlyn Zhou, Kawin Ethayarajh, Dallas Card, and Dan Jurafsky. 2022{\natexlab{a}}.
\newblock Problems with cosine as a measure of embedding similarity for high frequency words.
\newblock In \emph{Proceedings of the 60th Annual Meeting of the Association for Computational Linguistics (Volume 2: Short Papers)}, pages 401--423.

\bibitem[{Zhou et~al.(2022{\natexlab{b}})Zhou, Ethayarajh, Card, and Jurafsky}]{zhou-etal-2022-problems}
Kaitlyn Zhou, Kawin Ethayarajh, Dallas Card, and Dan Jurafsky. 2022{\natexlab{b}}.
\newblock \href {https://doi.org/10.18653/v1/2022.acl-short.45} {Problems with cosine as a measure of embedding similarity for high frequency words}.
\newblock In \emph{Proceedings of the 60th Annual Meeting of the Association for Computational Linguistics (Volume 2: Short Papers)}, pages 401--423, Dublin, Ireland. Association for Computational Linguistics.

\bibitem[{Zipf(1945)}]{zipf1945meaning}
George~Kingsley Zipf. 1945.
\newblock The meaning-frequency relationship of words.
\newblock \emph{The Journal of general psychology}, 33(2):251--256.

\end{thebibliography}

\appendix
\section{Additional Details}
\label{section:appendix_identity_probe}

Here, we provide additional details about the experimental set-up of \wc.

We used the logistic regression model from the scikit-learn library using a one-vs-rest (OvR) scheme.

\textit{Did you try other ways of creating embeddings?}
We explored alternative methods of creating word embeddings, such as various ways of concatenating layers, but they produced almost identical results. 

\textit{Did you perform any preprocessing?}
We filtered out short (<20 characters) and long (>512 characters) sentences, and matched keywords on token IDs to ensure punctuation and casing are consistent across examples.

\textit{Which hyperparameters did you use?}
Our task is also trained without any use of hyperparameters or special pre-processing steps to help address the concerns pointed out by \citet{liu-etal-2019-linguistic, hewitt-liang-2019-designing}.

\textit{How does this differ from BERT's training task and other works?}
The identity retrieval task differs from the masked LM training task: in masked LM training, the  word identity must be predicted from its \textbf{surrounding context} rather than the embedding itself. Our task is also related to but different from  the ``word identity'' classifier of \citet{zhang-bowman-2018-language} which predicts the identity of a \textbf{neighboring} word.

\textit{What about OOV words?}
\label{section:OOV}
For the error analysis, we used the embedding of the first subtoken. Throughout the rest of the paper, we average the subtokens following \citet{pilehvar-camacho-collados-2019-wic} and \citet{blevins-zettlemoyer-2020-moving}. Our decision to use the first subtoken in the error analysis section was to investigate the impacts of tokenization and perform analysis on token frequencies of the first subtokens when words were OOV.

\textit{In the benchmarking tasks, does your decision to represent a word via the embedding of its first token impact a word’s identifiability?} 
We find this is largely not the case. BERT-Base has a \textasciitilde30,000 token vocabulary, with words that occurred over \textasciitilde10,000 times in its original training data considered in the vocabulary. The word “intermission”, is out-of-vocabulary and is tokenized into “inter” and “\#\#mission”, and we would use the (extremely ambiguous) first token “inter” to represent “intermission”. 

\begin{figure}
\begin{subfigure}{0.48\columnwidth} 
\includegraphics[width=\textwidth]{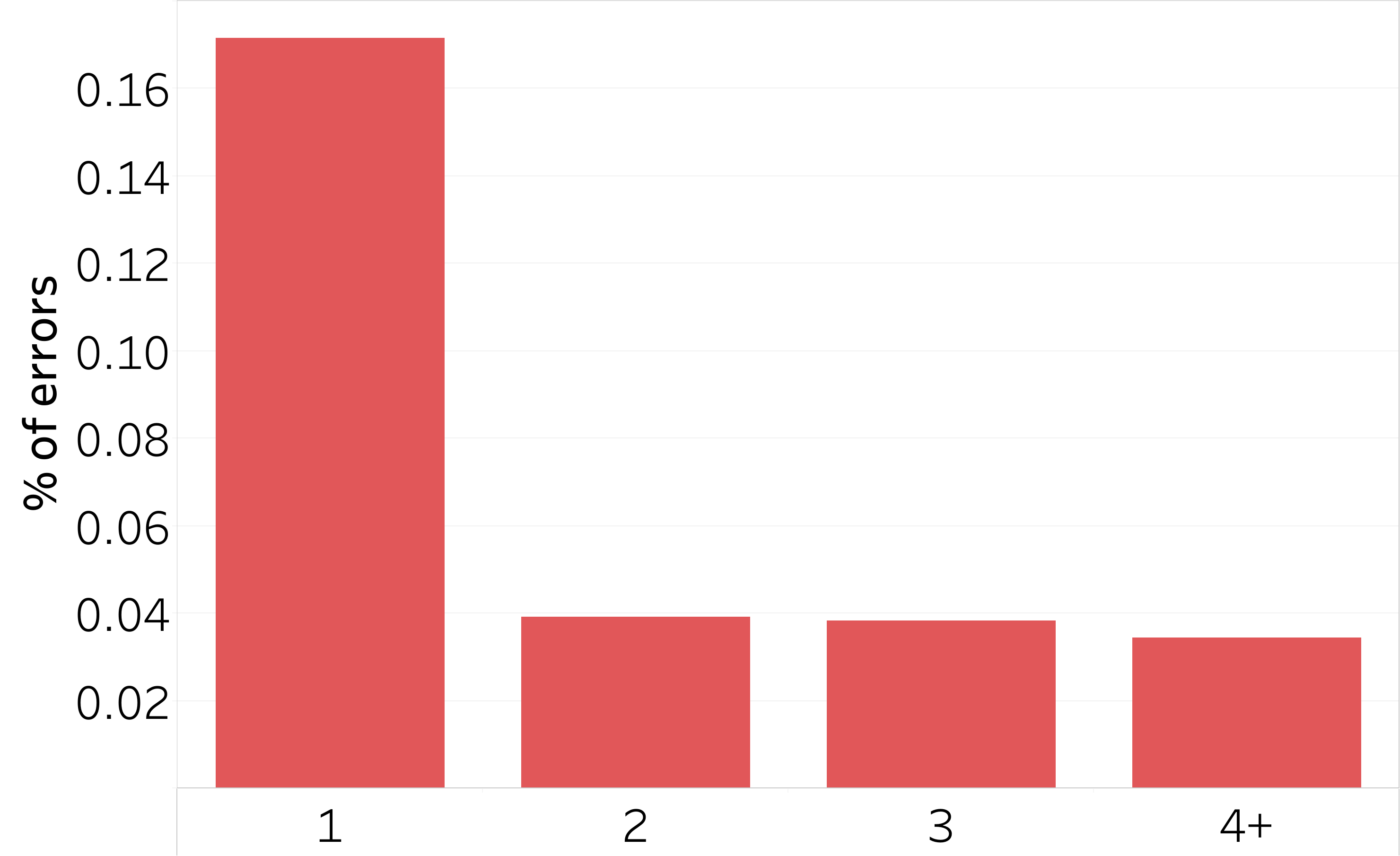} 
\caption{Tokens}
\label{fig:tokens}
\end{subfigure}  
\hfill 
\begin{subfigure}{0.48\columnwidth} 
\includegraphics[width=\textwidth]{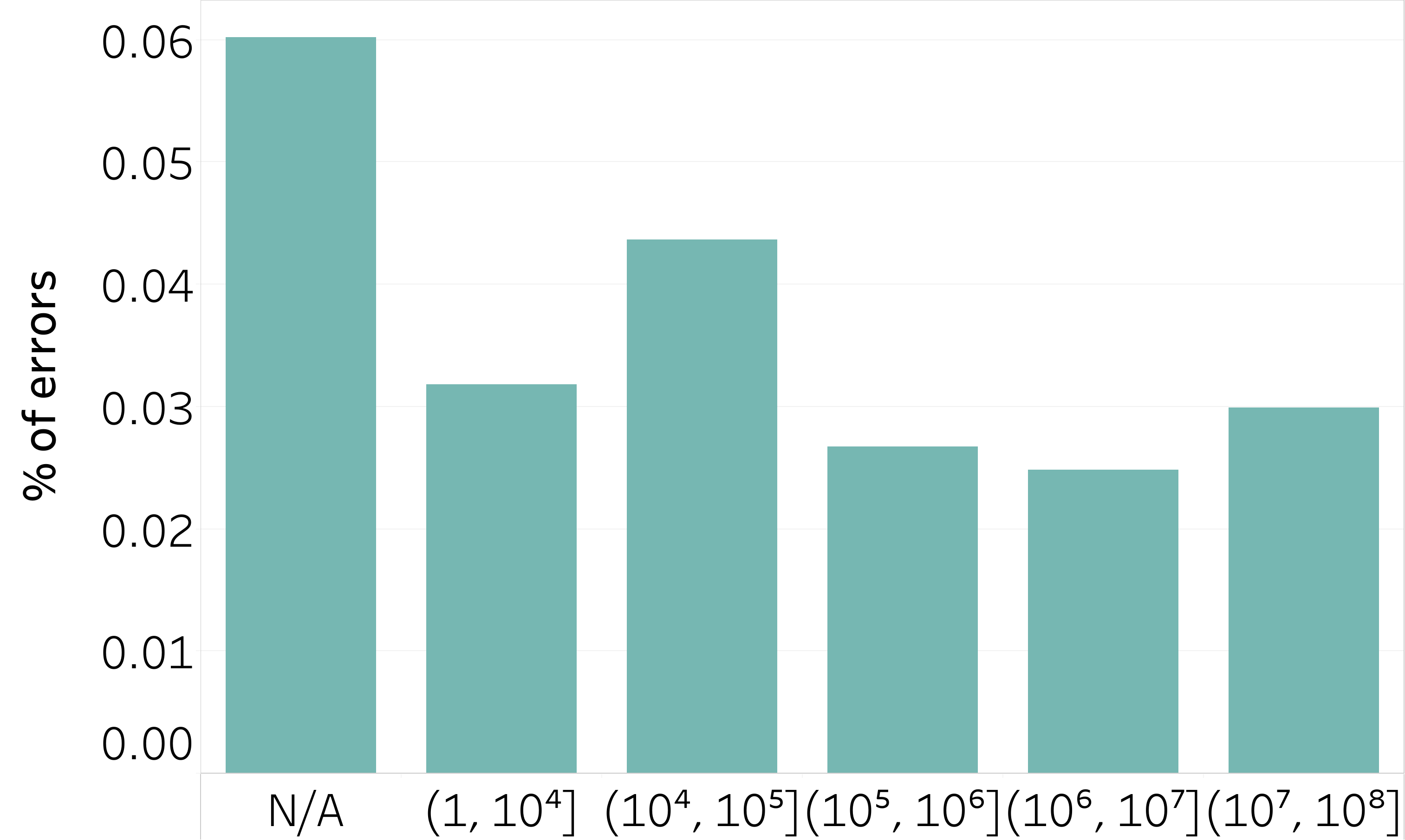} 
\caption{1st Token Freq}
\label{fig:first_token}
\end{subfigure}
\caption{The bar charts above highlight the percentage of errors for words binned by tokens and frequencies of the first subtoken for OOV words. (a) errors by number of tokens (b) errors by frequency of the first token} 
\label{fig:OOV_figure}
\end{figure}

Surprisingly, using only the first token to represent an OOV word had little impact on the identifiability of words, suggesting that these embeddings could capture enough context to differentiate themselves from words with identical prefixes. We find that words tokenized into multiple pieces had lower error rates (4\%) than words that remained whole (17\%) (see figure \ref{fig:tokens}). In other words, the words “intermission”, “interpromotional”, “interwar”, and “interwoven” are distinguishable from one another even though each is tokenized into “inter” and subsequent tokens and only the first token’s embedding is used. That is, the context (namely, the subsequent token “\#\#mission”) sufficiently changed the BERT embedding for “inter” to make it identifiable in context. The fact that single tokens words (which are in vocabulary and generally more frequent) performed worse as a group is likely explained by our prior finding that high frequency words have lower performance on this task (see figure \ref{fig:first_token}).

\subsection{Error Analysis}
Although \wc is relatively accurate (> 90\% accuracy), it can still makes mistakes, particularly with highly frequent or polysemous words.
%\footnote{Although not critical to this paper, we also include error analysis on the impacts of tokenization and OOV words in Appendix \ref{section:OOV}.}

\begin{figure}
  \begin{subfigure}{0.48\columnwidth}
  \includegraphics[width=\textwidth]{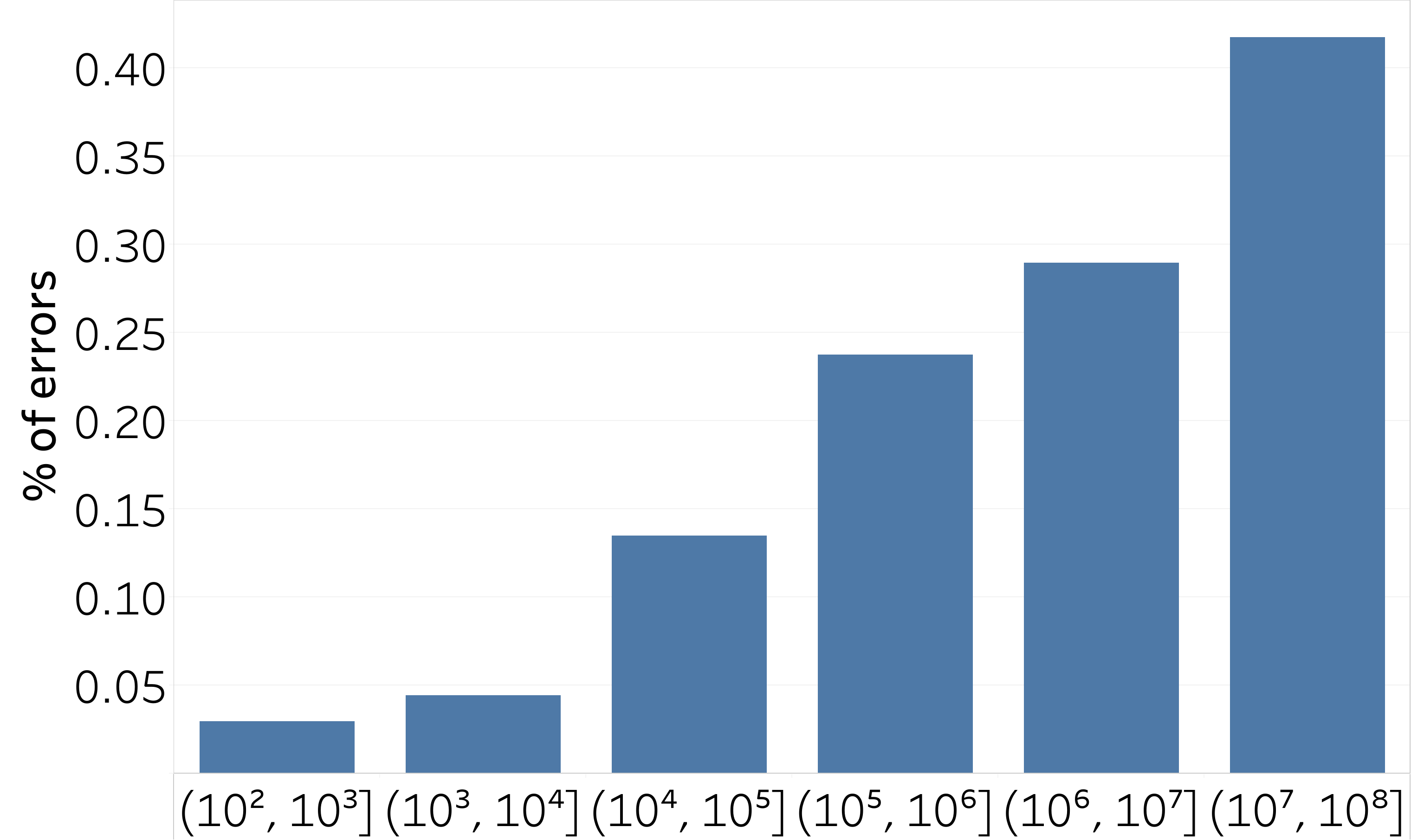}
  \caption{Frequency}
  \label{fig:frequency}
  \end{subfigure}
  \begin{subfigure}{0.48\columnwidth}
  \includegraphics[width=\textwidth]{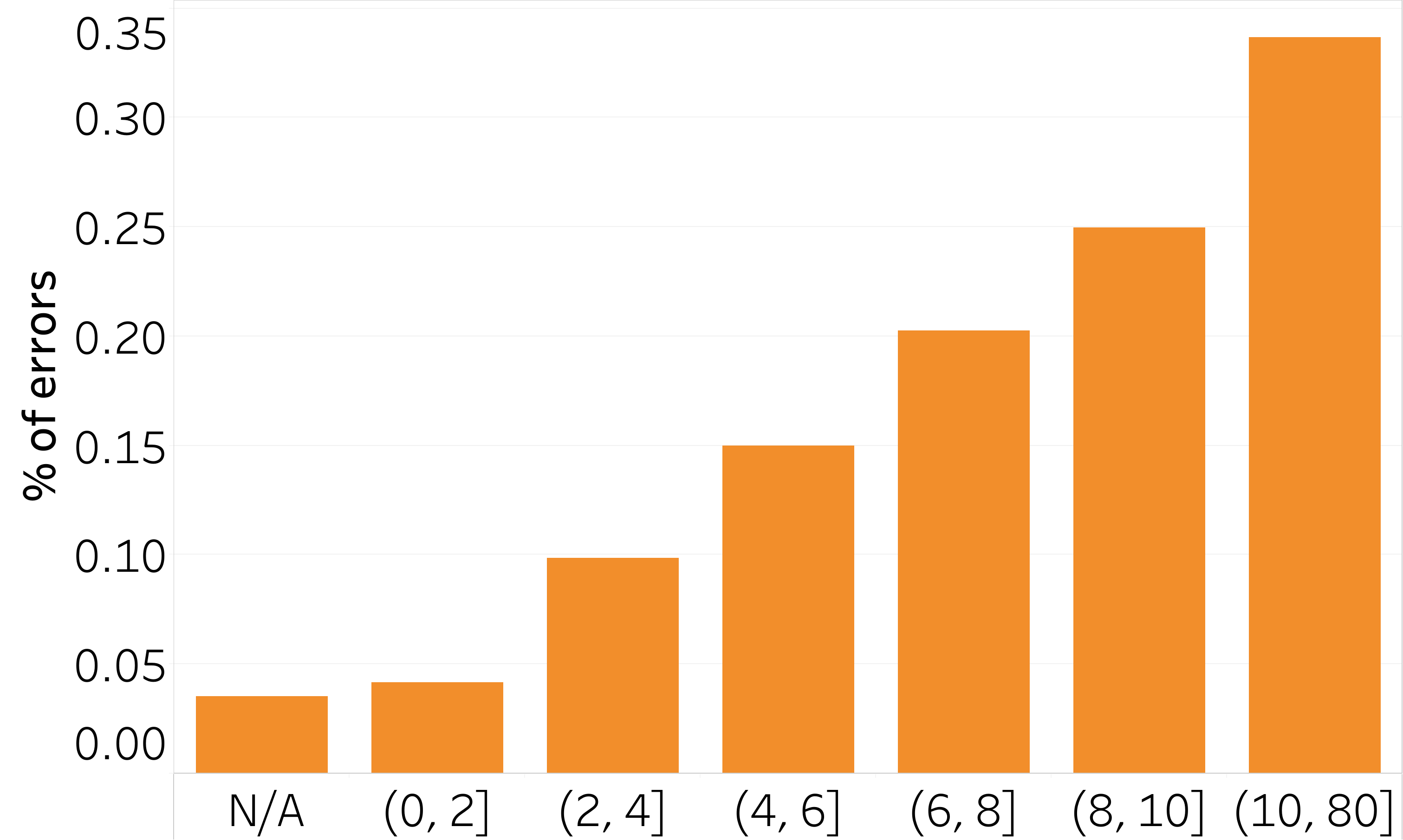}
  \caption{Senses} 
  \label{fig:senses}
  \end{subfigure} 
  \caption{The percentage of errors for words binned by frequency and number of senses.}
  \label{fig: radius vs word frequency}
\end{figure}

\paragraph{Frequency} We find that a word's training data frequency correlates negatively with identifiability. For example, the error rate of words with over 10 million training data occurrences is 42\%, compared to an error rate of 3\% for rare words with between 100 and 1000 training data occurrences.

\paragraph{Polysemy}
One explanation for the poor performance of high-frequency words could be the high polysemy of these words \citep{zipf1945meaning}. Indeed, \wc makes more errors with polysemous words. Very polysemous words (more than 10 senses in WordNet) are 8 times more likely than monosemous words to be misidentified (34\% versus 4\%, see figure \ref{fig:senses}). 

\paragraph{Geometric Space}
Another explanation for lower linear separability of high frequency words is that embeddings of high frequency words are typically more dispersed in geometric space than low frequency words \cite{zhou-etal-2022-problems}. This would most likely lead to difficulty in identifying them with a simple logistic regression model.

\section{Details and Full Results from Section \ref{sec:validation_experiments}}
\label{appendix:validation_details}
\paragraph{Implementation}
Out-of-vocabulary words here are represented as the average of the words' tokens, following \citet{pilehvar-camacho-collados-2019-wic} and \citet{blevins-zettlemoyer-2020-moving}.
We experiment with a variety of embedding methods, taking the last layer and taking the first subtoken of out-of-vocabulary words and find comparable results.

\paragraph{Similarity Experiments}
For cosine, we took 30 samples of each word and we took the average embedding (this is standard practice). For \wc, we again took 30 samples and we averaged the vectors of the predicted probabilities before taking the target probability values.

\paragraph{Feature Extraction Experiments}
Word sampling for target and seed words is done to speed up the computation, we did not find significant differences with different samples (nonetheless, having at least 1000 embeddings to train \wc is necessary to get good and stable results).

\paragraph{Models used:}
\begin{itemize}
    \item ``bert-base-cased"
    \item ``dbmdz/bert-base-italian-cased"
    \item ``dbmdz/bert-base-french-europeana-cased"
\end{itemize}

\subsection{Seed and Target Words Used}
\label{sec:seeds}
\textbf{Sentiment Classification} 
\begin{itemize}
    \item  \textbf{Task}: Classifying concepts based on sentiment by using the NRC corpus \cite{mohammad-etal-2013-nrc}. Target words: 98 positive and 98 negative words. Seed words: ``positive'' and ``negative''.
    \item \textbf{Corpus}: wikitext-103-v1 from HuggingFace. We remove sentences that are shorter than 15 tokens and longer than 200 tokens.
    \item \textbf{Sampling}: We sample 1000 occurrences of ``positive'' and 1000 occurrences of ``negative''. For each target word, we sample 30 occurrences.
\end{itemize}

\textbf{Grammatical Gender in French and Italian} 

Experiment 1: 
\begin{itemize}
    \item \textbf{Task}: Classifying 
concepts by the grammatical gender of nouns. 
\item \textbf{Corpus}: Latest Italian Wikipedia abstracts from DBPedia. We removed sentences shorter than 20 tokens and longer than 100 tokens.
\item \textbf{Sampling}: Target words: 140 Italian nouns. Seed words: 59 Italian masculine and feminine adjectives. For each target word, we sample 30 occurrences. For each seed word, we sample 20 occurrences. Seed and target words have been filtered with respect to frequency. Data comes from Flex-IT~\cite{pescuma2021form}.
\end{itemize}

Experiment 2: 

\begin{itemize}
    \item \textbf{Task}: Classifying 
concepts by the grammatical gender of nouns.
\item \textbf{Corpus}: Latest French Wikipedia abstracts from DBPedia. We removed sentences shorter than 20 tokens and longer than 100 tokens.
\item \textbf{Sampling}: Target words: 201 French nouns. Seed words: 65 French masculine and feminine adjectives. Seed and target words have been filtered with respect to frequency. Data comes form Lexique383~\cite{npbf04}.
\end{itemize}

\textbf{BERT Concept Net Classification Land-Sea} 
\begin{itemize}
    \item \textbf{Task}: Classifying concepts by classes based on the ConceptNet dataset \cite{dalvi2022discovering}, predicting if an animal is a sea or land animal.
    \item \textbf{Corpus}: wikitext-103-v1 from HuggingFace. We remove sentences that are shorter than 15 tokens and longer than 200 tokens.
    \item \textbf{Sampling}: Target words: 64 land or sea animals. Seed words: category names: ``land'' and ``sea''. We sample 1000 occurrences of each seed word. For each target word, we sample 30 occurrences.
\end{itemize}

\textbf{BERT Concept Net Classification Fashion-Gaming} 

\begin{itemize}
    \item Task: Classifying concepts by classes based on the ConceptNet dataset \cite{dalvi2022discovering}, predicting if a concept comes from the fashion domain or the design domain.
    \item Corpus: wikitext-103-v1 from HuggingFace. We remove sentences that are shorter than 15 tokens and longer than 200 tokens.
    \item Sampling: Target words: 29 terms related to fashion or gaming. Seed words: category names: ``fashion, clothes'' and ``gaming, games''. We sample 500 occurrences of each seed word. For each target word, we sample 30 occurrences.
\end{itemize}

% \label{appendix:cosine_results}
% \begin{table*}[]
%     \centering
%     \resizebox{\textwidth}{!}{%
%     \begin{tabular}{lccccc} \toprule
%     \textbf{Experiment} & \textbf{\wc} & \textbf{Cosine 1} & \textbf{Cosine 2} & \textbf{Cosine 3} & \textbf{Ave. Cosine} \\ \midrule
%      Sentiment Classification & \textbf{0.79} & 0.75 & 0.71 & 0.84 & 0.73 \\
%      Grammatical Gender (Italian) & \textbf{0.93} & 0.80 & 0.80 & 0.71 & 0.77 \\
%      Grammatical Gender (French) & 0.85 & \textbf{0.86} & 0.86 & 0.83 & 0.85 \\
%      ConceptNet Domain (Fashion-Gaming) & 0.90 & \textbf{0.93} & 0.93 & 0.90 & 0.92 \\
%      ConceptNet Domain (Sea-Land Animals) & \textbf{0.83} & 0.79 & 0.80 & 0.61 & 0.73 \\
%      \midrule
%      Average & \textbf{0.86} & 0.83 & 0.82 & 0.76 & 0.80 \\ \bottomrule
%     \end{tabular}
%     }
%     \caption{Full results from Section \ref{sec:validation_experiments}. We compare the results of \wc to cosine similarity which we operationalize in one of three ways: we measure cosine similarity in one of three ways 1) the distance between the centroids of the seed words and the target words 2) the average distance each of the target word to the centroid of the seed words 3) the average distance of each target word to each seed word (no centroids)}
%     \label{tab:cosine_results}
% \end{table*}

\section{Capturing Trends in Inflation}
\label{sec:finance_experiment}

In a very preliminary experiment, we also apply {\wc} to a novel social science domain:  representation of economical value or financial meaning. Here we test whether we can recover the financial value of goods from their embeddings and use them to predict changes in those values  -- inflation. We choose inflation since it is easy to quantify and explores a novel domain for this sort of computational meaning. However, the results are preliminary, these trends are extremely complex, and more diverse and domain-specific data could help improve our understanding of applications to this domain.    

We used the California Digital Newspaper Collection (CDNC)\footnote{\url{https://cdnc.ucr.edu/}}, a newspaper corpus that covers the years 1846-2023. We segmented the data into temporal periods based on trends in the Dow Jones Index (DJI)\footnote{\url{https://www.macrotrends.net/1319/dow-jones-100-year-historical-chart}}, aggregating intervals that exhibited the same index fluctuation directions. At the end of the process, we had 17 different data segments, spanning the years 1915-2009. We then further trained the last layer of a 12-layer BERT model for each temporal segment, to create embeddings that capture a particular historical period, with the goal of capturing the temporal change in the value of money.

To quantify the change in the value of money, we trained \wc for every temporal segment of the data. Its goal was to map from the contextual embedding of the `` \$'' token to the (bucketed) monetary value that accompanied that dollar sign. Thus, for each temporal segment, we extract all sentences containing ``\$'', and use the contextual embedding of \$ for predicting the bucketed monetary value from the original sentence. For example, if the sentence is ``The price of gas increased to \$3 per gallon!'', we train a linear regression model to correctly map the \$ embedding to the bucket that contains 3.\footnote{The average correlation coefficient of the trained \wc regressors across the different temporal segments is 0.790, indicating a strong correlation between the  \$ embeddings and their numerical values in context.}

We used all of the temporal \wc classifiers to predict the monetary values of items in a typical basket of goods (e.g., egg, milk, gasoline, car, etc)\footnote{To make the analysis as similar to the real CPI as possible, we used the reported products from the website of the U.S. Bureau of labor statistics, keeping only products that were found in all segments (to avoid biasing our results by using products that were not invented in the past).}.
We then compare these predictions with two measures -- the historical Consumer Product Index (CPI) and the Dow Jones Index (DJI).

\begin{figure}[h!]
    \includegraphics[width=0.48\textwidth]{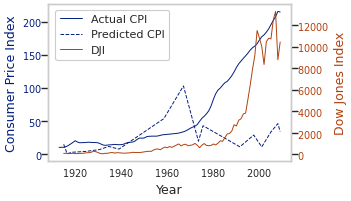}
    \caption{Average CPI, DJI, and \wc values between the years 1915-2009. 
    For each temporal segment, the \wc values were calculated using the mean predicted value for each item in the basket of goods. We can see that until the 1970s \wc values followed the increasing CPI trend, but then dropped.  This could be a problem in our method, or could be caused by changes in the training text itself at that period of time, in any case  require further investigation that includes domain experts.}
    \label{fig:finance}
\end{figure}

The correlation between CPI and DJI, is very high (0.966), indicating they capture similar trends. The correlations of \wc values with CPI (0.187) and DJI (0.169) are positive and significant but low. This low correlation indicates that inflation prediction is a complicated task, which \wc gives us only a very partial window on; the weakness of fit is clear in inspecting Figure \ref{fig:finance}.
While this particular application of our measure is thus inconclusive, the results do suggest that further study involving domain experts could be instructive on whether \wc or similar methods could be used to study financial values in text. 

\section{Details and Full Results from Section \ref{sec:finance_experiment}}
\label{appendix:financial_report}
\paragraph{Data Segmentation}
We segment the temporal data based on the Dow Jones Index  trend\footnote{\url{https://www.macrotrends.net/1319/dow-jones-100-year-historical-chart}} and aggregate intervals with the same fluctuation directions (see Table \ref{tab:DJITrends}).
\begin{table*}[h!]
    \centering
    \begin{tabular}{lc} \toprule
    \textbf{Year} & \textbf{DJI Avg. Annual Change} \\ \midrule
      1915 & 81.49\% \\ 
 1916-1917 & -12.95\%  \\
 1918-1919 & 20.48\%  \\
 1921-1928 & 20.48\% \\
 1929-1932 & -31.67\%  \\
 1933-1936 & 30.02\%  \\ 
 1937-1941 & -7.16\%  \\ 
 1956-1961 & 9.97\%  \\ 
 1962-1972 & 3.86\%  \\ 
 1973-1974 & -22.08\%  \\
 1975-1976 & 12.35\%  \\
 1988-1995 & 13.53\%  \\
 1996-1999 & 22.49\%  \\
 2000-2002 & -10.01\%  \\
 2003-2007 & 11.04\% \\
 2008 & -33.84\% \\
 2009 & 18.82\%  \\ 
 \bottomrule
    \end{tabular}
    \caption{Years aggregated by DJI fluctuation directions}
    \label{tab:DJITrends}
\end{table*}

\paragraph{Data Pre-processing}
We use California Digital Newspaper Collection \cite{CDNC2024data} spanning from 1915 to 2008. The data is pre-processed in the following manner for model continual training:
\begin{itemize}
    \item Convert all text to lowercase.
    \item Remove low-quality text corpuses, defined as those where more than 20\% of the characters are non-alphanumeric symbols or where more than 20\% of words are highly segmented (a single word tokenized into more than two segments), due to poor optical character recognition from scans of historical documents. 
    \item The dataset of each training segment has 10,240 training documents, 1280 test documents and 1280 validation documents, each containing an average of 350 tokens.
\end{itemize}

\paragraph{Continual Training}
We fine-tune the last layer of the 12-layer bert-base-uncased model, which comprises 7,087,872 trainable parameters. We use a learning rate of $2 \times 10^{-5}$ and a weight decay of 0.01. Each model takes 3 hours to fine-tune with Google Cloud T4 GPUs.\footnote{\url{https://cloud.google.com/compute/docs/gpus\#t4-gpus}}. 
\paragraph{Training \wc}
We extract 2,000 occurrences of the "\$" token from each segment. Each token is part of a 128-character window and must be followed by a numeric value. We get the contextualized embedding of the tokens using the fine-tuned models and bucketize the 2000 numeric values into 60 buckets to reduce noise in the data. We then train a linear regression for each time segment.

\paragraph{Calculating CPI}
To calculate the Consumer Price Index (CPI), we construct a basket of goods consisting of the following items: \{"car", "rent", "hat", "wine", "jewelry", "shirt", "chicken", "milk", "furniture", "egg", "shoe", "pork", "gasoline", "beef", "coffee", "bus"\}. We identify occurrences of the "\$" token that are followed by a numeric value and keep those where terms from our basket of goods appear within a 20-word window. The numeric values are then masked, and the trained \wc classifier is used to predict the value associated with each "\$" token.

\paragraph{Models used:}
\begin{itemize}
    \item ``bert-base-uncased"
\end{itemize}

\paragraph{Rate of change in CPI, DJI, and \wc values:}
Rate of change in \wc values compared with the rate of change in CPI and DJI values (the mean annual change in values per temporal segment). The correlation between the change in CPI and DJI values is almost zero (-.006), suggesting they capture quite different trends. The correlation of CPI change and \wc change is negative (-0.226), and the correlation between the changes in DJI and \wc values is positive and significant (0.387).

\end{document}